\documentclass[journal,twoside,web]{ieeecolor}
\usepackage{subfigure}
\usepackage{color}
\usepackage{arydshln} 
\usepackage{generic}
\usepackage{cite}
\usepackage{amsmath,amssymb,amsfonts}
\usepackage{algorithmic}
\usepackage{graphicx}
\usepackage{algorithm,algorithmic}
\usepackage{hyperref}
\usepackage{multirow}
\hypersetup{hidelinks=true}
\usepackage{textcomp}

\DeclareUnicodeCharacter{FF5E}{\textasciitilde}

\def\BibTeX{{\rm B\kern-.05em{\sc i\kern-.025em b}\kern-.08em
    T\kern-.1667em\lower.7ex\hbox{E}\kern-.125emX}}
\begin{document}
\epstopdfsetup{outdir=./}
\title{Masked Contrastive Reconstruction for Cross-modal Medical Image-Report Retrieval}
\author{Zeqiang Wei, Kai Jin, and Xiuzhuang Zhou, \IEEEmembership{Member, IEEE} 
\thanks{Zeqiang Wei, Kai Jin and Xiuzhuang Zhou are with the School of Artificial Intelligence, Beijing University of Posts and Telecommunications, Beijing 100876, China (e-mail: weizeqiang@bupt.edu.cn).}}

\maketitle

\begin{abstract}
Cross-modal medical image-report retrieval task plays a significant role in clinical diagnosis and various medical generative tasks. 
Eliminating heterogeneity between different modalities to enhance semantic consistency is the key challenge of this task. 
The current Vision-Language Pretraining (VLP) models, with cross-modal contrastive learning and masked reconstruction as joint training tasks, can effectively enhance the performance of cross-modal retrieval.
This framework typically employs dual-stream inputs, using unmasked data for cross-modal contrastive learning and masked data for reconstruction. 
However, due to task competition and information interference caused by significant differences between the inputs of the two proxy tasks, the effectiveness of representation learning for intra-modal and cross-modal features is limited.
In this paper, we propose an efficient VLP framework named Masked Contrastive and Reconstruction (MCR), which takes masked data as the sole input for both tasks.
This enhances task connections, reducing information interference and competition between them, while also substantially decreasing the required GPU memory and training time.
Moreover, we introduce a new modality alignment strategy named Mapping before Aggregation (MbA). 
Unlike previous methods, MbA maps different modalities to a common feature space before conducting local feature aggregation, thereby reducing the loss of fine-grained semantic information necessary for improved modality alignment.
Qualitative and quantitative experiments conducted on the MIMIC-CXR dataset validate the effectiveness of our approach, demonstrating state-of-the-art performance in medical cross-modal retrieval tasks.
\end{abstract}


\section{Introduction}
\label{sec:introduction}

\begin{figure}[!t]
    \centering
    \subfigure[The VLP framework takes as input both masked and unmasked data]{
        \includegraphics[width=1.0\columnwidth]{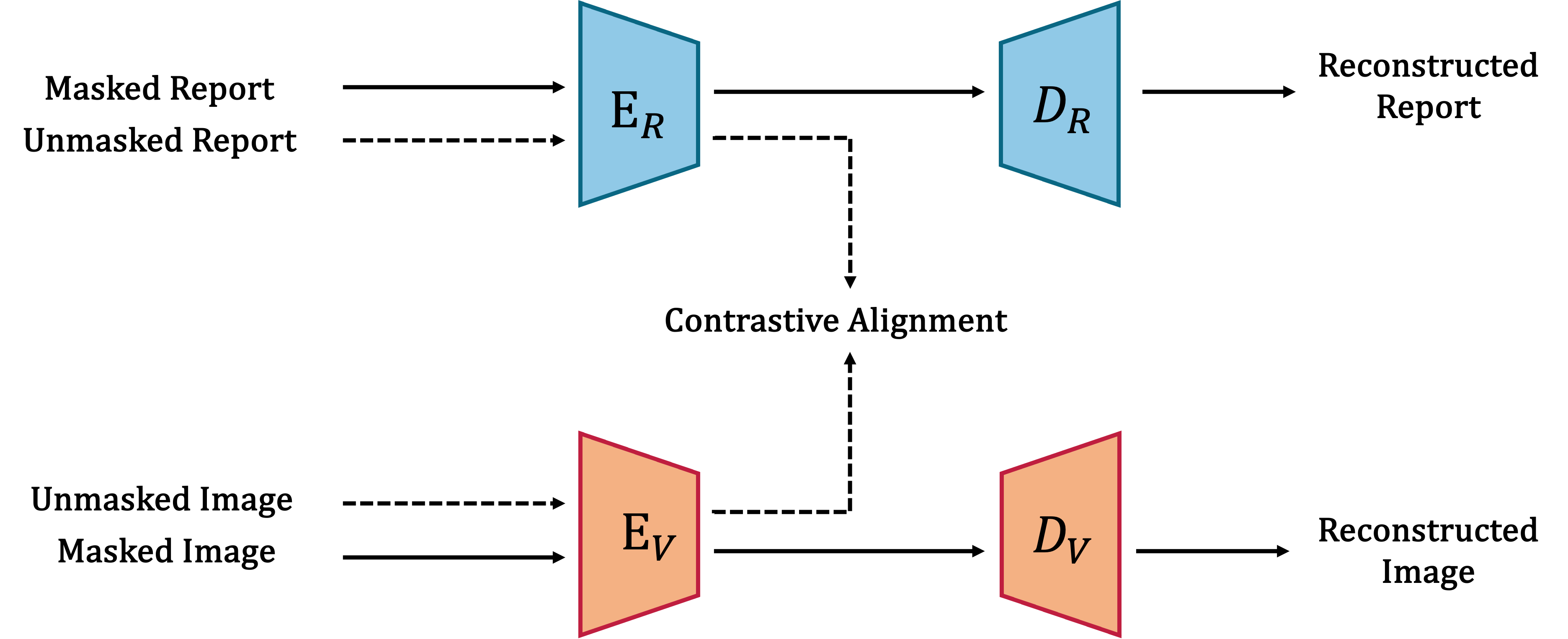}
        \label{double_framework}} \\
    \noindent\rule{8.6cm}{0.5pt} \\
    \subfigure[Our MCR framework takes as input masked data alone]{
        \includegraphics[width=1.0\columnwidth]{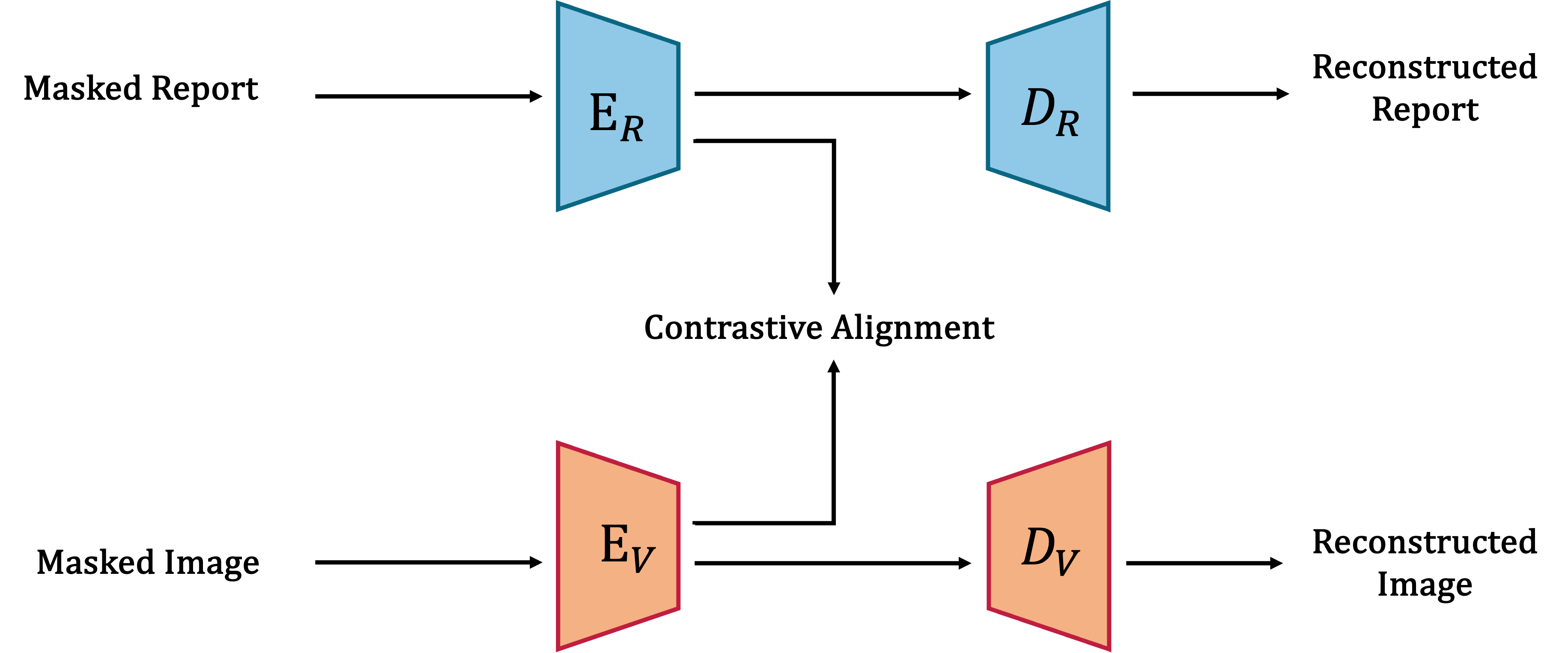}
        \label{alone_framework}}
    \caption{The VLP framework (a) takes as input both masked and unmasked data, and uses unmasked features for cross-modal contrastive learning and masked features for reconstruction. Our MCR framework (b) takes as input masked data alone, and uses masked features for both cross-modal contrastive learning and masked reconstruction.}
    \label{framework}
    \vspace{-0.7cm}
\end{figure}

\IEEEPARstart{T}{he} \textbf{C}ross-modal \textbf{M}edical \textbf{I}mage-\textbf{R}eport \textbf{R}etrieval task (CMIRR), from a large-scale database, another modality data that is semantically similar to an chest X-ray image or radiology report used as a query. In contrast to queries that only depend on images, cross-modal retrieval enables healthcare professionals to utilize specialized anatomical and clinical pathological knowledge during the retrieval process. This enhances the interpretability and credibility of search results, making them more intuitive and understandable. CMIRR is not only beneficial for clinical diagnosis, but also widely applied in various medical generative tasks such as report generation, text-image synthesis, and visual question answering. As deep learning technology advances, cross-modal retrieval algorithms have evolved from traditional methods such as \textbf{C}anonical \textbf{C}orrelation \textbf{A}nalysis (CCA) and autoencoders to CNN-RNN, Transformer, \textbf{V}ision-\textbf{L}anguage \textbf{P}re-training (VLP), and cross-modal generation methods. Among them, the robust modal feature extraction of VLP significantly advances cross-modal retrieval, establishing it as one of the most advanced and promising algorithms. This paper focuses on effectively leveraging the VLP algorithm to further enhance CMIRR performance.

Cross-modal contrastive learning and masked reconstruction are the two most effective pre-training proxy tasks within the VLP. The existing VLP framework that combines cross-modal contrastive and masked reconstruction commonly takes pairs of unmasked and masked data as inputs \cite{zhang2023multi, Ji_2023_CVPR, dong2023maskclip}. It utilizes a shared modality encoder to generate unmasked and masked features. Subsequently, it uses the unmasked features for cross-modal contrastive learning and the masked features for masked reconstruction learning, as depicted in Fig.\ref{double_framework}. 
Due to significant differences between the inputs of these proxy tasks in this framework, it leads to task competition and information interference during training, severely limiting the learning of intra-modal features and cross-modal common semantic features, as shown in \cite{shrestha2023medical}.
For this, we propose an efficient VLP framework called \textbf{M}asked \textbf{C}ontrastive and \textbf{R}econstruction (MCR), which utilizes masked features for both cross-modal contrastive learning and masked reconstruction, as illustrated in Fig.\ref{alone_framework}. 
By imposing cross-modal semantic consistency constraints on masked features, we further enhance the extraction of common semantic features across modalities while improving the fine-grained feature extraction capacity of the modal encoder. This contributes to enhancing cross-modal alignment and improving the performance of cross-modal retrieval tasks. The subsequent experimental results in this paper demonstrate the effectiveness of the MCR framework. Without differing data distributions between the inputs of both proxy tasks, the MCR framework integrates these tasks more effectively. 
{\color{black}{Moreover, as it no requires unmasked data input, MCR offers advantages in reducing memory usage and enhancing training speed compared to existing frameworks. 
For example, our MCR framework conserves 75\% of GPU memory and reduces time consumption by 50\% during training.}}

The core challenge in cross-modal retrieval tasks lies in mitigating the heterogeneity between multiple modalities to measure the semantic similarity between image and text data. Currently, the most common solutions are cross-modal alignment and multi-modal fusion. Cross-modal alignment involves mapping features from different modalities into a shared feature space to compute similarity between distinct modal features within that space. Multi-modal fusion refers to merging two different modality features into a new feature, using this new feature to calculate the matching score between the two modality features. Due to the substantial computational burden imposed by multi-modal fusion, which renders it impractical for large-scale medical retrieval applications, this paper primarily focuses on cross-modal alignment solutions.

In general scenarios, cross-modal alignment commonly aggregates local features within different modal spaces to obtain global features. Then, these global features are mapped to a common space for alignment learning. We refer to this alignment strategy as \textbf{A}ggregate \textbf{b}efore \textbf{M}apping (AbM). Unlike general image captions, radiology reports typically provide detailed descriptions of various regions within chest X-ray images, lacking common sequential or causal associations between these descriptions. Additionally, the regions describing lesions in images corresponding to the reports are often narrowly distributed and sparse. So, similar to general scenarios, aggregating local features within the original modality space often results in different modal consist information loss. 
Although this missing is inevitable during local feature aggregation, it is essential for medical cross-modal alignment.
As this missing occurs prior to mapping into a common alignment space, it leads to a suboptimal learned alignment space. 
Deep learning tend to focus on learning simpler content, which might exacerbate the learning of suboptimal alignment spaces, leading to a narrower and more one-sided focus on modal features. For this, we propose a method that initially maps all features from the original modality into a common space, aiming to minimize the loss of modal information. Subsequently, in this common space, these features are aggregated into global features for alignment learning. This approach enables a more refine alignment learning by preserving finer details. We refer to this alignment strategy as \textbf{M}apping \textbf{b}efore \textbf{A}ggregate (MbA). The experiments provided quantitative and qualitative evidence of the superiority of the MbA alignment strategy.

In summary, the contributions of this paper can be summarized as: 
\begin{enumerate}
    \item This paper introduces an efficient VLP framework called MCR, which uniquely conducts contrastive learning and masked reconstruction solely using masked features.
    This strengthens the connections between various tasks, reducing interference and competition among them, thereby enhancing the extraction of intra-modal features and cross-modal common semantic features. 
    Furthermore, benefiting from requiring only masked inputs, MCR provides advantages in reducing GPU memory usage and accelerating training speed.
    For example, our MCR framework conserves 75\% of GPU memory and reduces time consumption by 50\% during training.
    \item Based on the characteristics of Chest X-ray images and radiology reports, this paper discovers that certain features crucial for cross-modal consistency are lost during the aggregation process before alignment. To address this, we introduce an alignment strategy called MbA, aiming to prevent significant information loss and achieve a more refined cross-modal alignment.
    \item We validated the effectiveness and superiority of the MCR framework and MbA strategy on the MIMI-CXR dataset, achieving the current state-of-the-art (SOTA) level in cross-modal retrieval.
\end{enumerate}

\section{Related Work}
\subsection{Cross-modal Retrieval}

The cross-modal alignment and multi-modal fusion are the two major solutions for cross-modal retrieval tasks. In cross-modal alignment, learning a common feature space is the core focus of research. In early cross-modal retrieval, employing \textbf{C}anonical \textbf{C}orrelation \textbf{A}nalysis (CCA) \cite{hardoon2004canonical} to build a shared space was the primary method to address cross-modal alignment, aiming to maximize the co-occurrence correlation between different modal features \cite{rasiwasia2010new, chen2012continuum, andrew2013deep, gong2014improving, klein2014fisher}. Following that, researchers proposed a series of methods based on contrast \cite{liu2023efficient}, triplet \cite{frome2013devise}, and ranking \cite{zhang2016pl} metric losses, aiming to better learn the cross-modal common feature space. Building on this, VSE++ \cite{faghri2018vse++} utilizes hard sample mining, PCME \cite{chun2021pcme} transforms deterministic features into probabilistic ones, and DiVE \cite{kim2023improving} introduces discrete prompts to enhance the extraction of modal features, further strengthening the alignment capability and robustness of the common space. For multi-modal fusion, the cross-modal attention mechanism has always been a key focus of research. SCAN \cite{lee2018stacked}, CAAN \cite{zhang2020context} , and other methods design various cross-modal attention networks, leveraging crucial context information from different modalities and computing fused features to determine image-text matching scores. Additionally, subsequent studies aimed to enhance the representation capability of fused features by incorporating new modules with different functionalities. For instance, CAMP \cite{wang2019camp} introduced an adaptive gating module to manage cross-modal information transmission and reduce the impact of negative samples and noise on the fused features. Thanks to the powerful modality feature extraction capability of VLP models, it significantly advances the development of cross-modal retrieval tasks. Methods like CLIP \cite{radford2021learning} and BLIP2 \cite{li2023blip} have become the most advanced and promising cross-modal retrieval methods currently available.

\subsection{Medical VLP}
According to \cite{park2023self}, contrastive learning is adept at capturing low-frequency global features, while mask reconstruction excels in extracting high-frequency local features. Combining these two pre-training tasks effectively enhances feature representation, leading to their widespread joint adoption as the primary pre-training proxy task for current VLP models. MPMA \cite{zhang2023multi} represents the first such VLP model in the medical field. It simultaneously inputs unmasked and masked data, obtaining unmasked and masked features through a shared modality encoder. Subsequently, it utilizes unmasked features for cross-modal contrastive learning and masked features for mask reconstruction learning, as illustrated in Fig.\ref{double_framework}. CMITM \cite{chen2023contrastive} retains a dual-input data pipeline in the text modality; however, in the image modality, it employs reconstructed image features for cross-modal contrastive learning, replacing unmasked image features and further promoting the integration of these two proxy tasks. Maco \cite{huang2023enhancing} is the closest to our proposed MCR framework, solely emphasizing masked contrastive reconstruction in the image modality while overlooking the text modality. Our subsequent ablation experiments highlight the necessity and superiority of conducting masked contrastive reconstruction simultaneously in both image and text modalities.

\subsection{Medical VLP for Cross-modal Retrieval}
In the general domain, the effectiveness of VLP models in cross-modal retrieval tasks has been widely demonstrated \cite{liu2022image, lu2022cots, wang2023agree}. This paper focuses on the application of medical VLP models in cross-modal retrieval tasks between chest X-ray images and radiology reports. The VLP model can be divided into single-stream and dual-stream architectures, corresponding respectively to multi-modal fusion and cross-modal alignment in cross-modal retrieval tasks.

The single-stream VLP architectures, corresponding to multi-modal fusion, combine multiple modalities to generate a unified representation of multi-modal features. They utilize calculated cross-modal feature matching scores for retrieval tasks.
CAMP pioneered modeling cross-modal feature matching as a binary classification: it classifies positively when an image and text are from the same pair, and negatively otherwise. MedVILL \cite{moon2022multi} concatenates image and text modal features as input to the cross-modal attention module, introducing the Sequence-to-Sequence attention mask for cross-modal information fusion. Additionally, it utilizes the \textbf{[CLS]} token as the multimodal representation for cross-modal feature matching computations. M3AE \cite{chen2022multi} introduces a multi-modal masked reconstruction task, applying masking operations to image-report data, leveraging cross-modal attention mechanisms to merge and complement information across different modalities, achieving enhanced cross-modal information fusion. Due to its capability in extracting fine-grained modal features through the masked reconstruction task, M3AE achieved state-of-the-art results in multiple downstream tasks. However, the high computational cost of multi-modal fusion, requiring queries to merge features with the entire database, significantly impacts its application in retrieval tasks. Thus, this paper primarily focuses on employing the dual-stream VLP architectures for retrieval tasks.

\begin{figure*}[!t]
\vspace{-15pt}
\centering
\includegraphics[width=18cm]{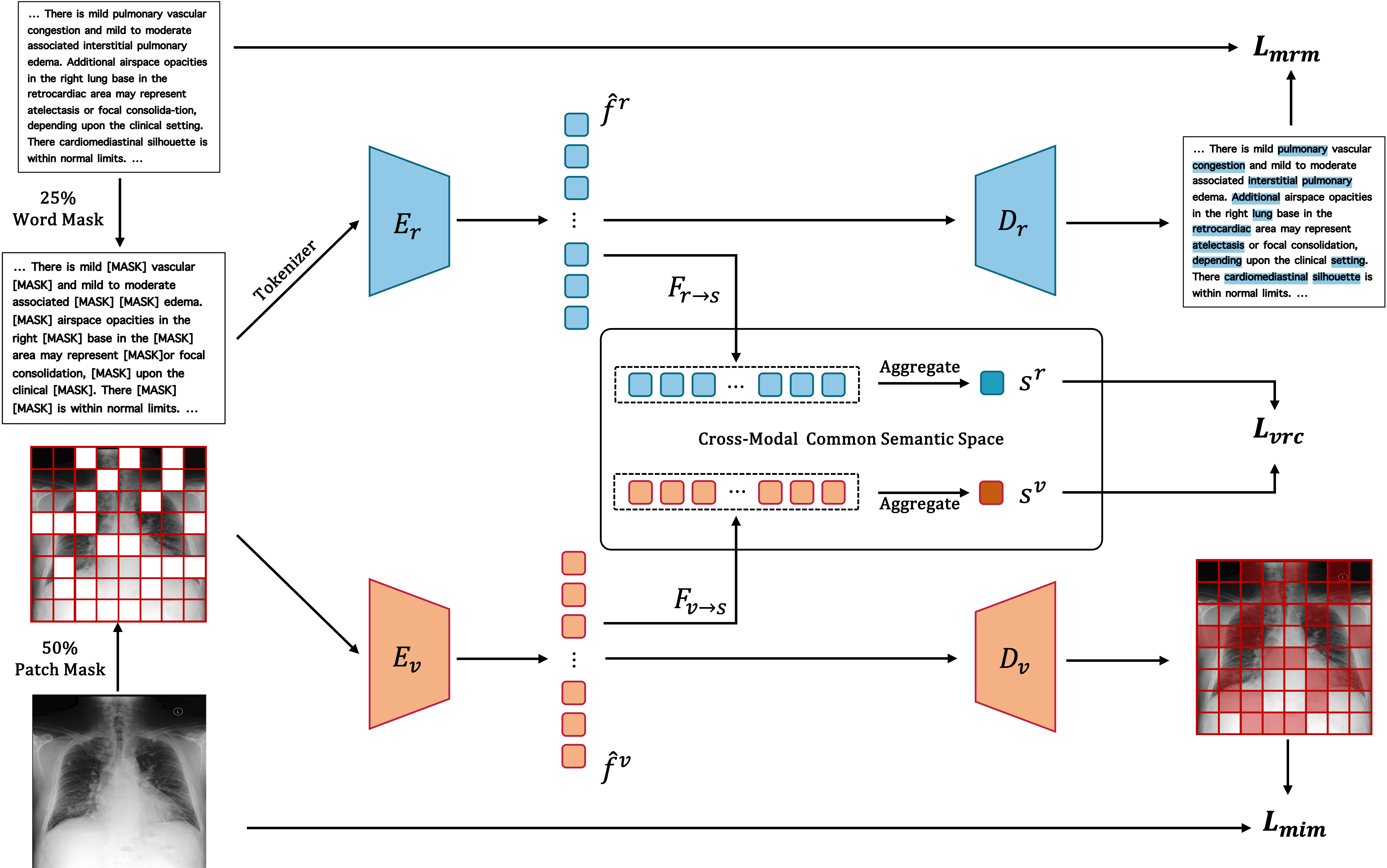}

\caption{
The overall pipeline of the MCR framework. It takes as input the masked chest X-ray images and reports with mask rates of 50\% and 25\%, respectively. 
Masked image features $\hat{f}^v$ and Masked report features $\hat{f}^r$ are extracted using the image encoder $E_v$ and text encoder $E_r$, respectively. 
Subsequently, $\hat{f}^v$ and $\hat{f}^r$ are mapped into a common semantic space using $F_{v \to s}$ and $F_{r \to s}$, followed by feature aggregation to obtain modality-aligned features $S^v$ and $S^r$. 
Additionally, $\hat{f}^v$ and $\hat{f}^r$ are used to reconstruct the original inputs by $D_v$ and $D_r$, respectively. 
$\mathcal{L}_{mim}$ and $\mathcal{L}_{mrm}$ respectively denote the masked reconstruction loss for chest X-ray image and corresponding report, and $\mathcal{L}_{vrc}$ denotes the loss of cross-modal consistency.}
\label{homepage}
\end{figure*}

The dual-stream VLP architectures correspond to cross-modal alignment. Images and text independently extract modal features, then learn and map them into a common feature subspace. Cross-modal retrieval tasks are accomplished by computing the similarity between different modal features within this common subspace. ConVIRT \cite{zhang2022contrastive} pioneered the use of the InfoNCE loss function \cite{oord2018representation} to align global features between images and text in medical VLP, profoundly influencing the development of numerous subsequent VLP models in the general domain. ConVIRT solely employs global features from image-text for alignment learning, while subsequent research attempts to enhance it by aligning cross-modal local, fine-grained features. GLoRIA \cite{huang2021gloria} achieves local feature alignment between image regions and report words through a weighted cross-modal attention mechanism. LIMITR \cite{Dawidowicz_2023_ICCV} further points out that not all local features contain crucial semantic information. They proposed a method similar to attention-pooling to learn the weights between local and global features and applied it in the alignment loss function. LoVT \cite{muller2022joint} posits that sentences, compared to words, more accurately express local semantic information in reports. Hence, they redefined the granularity of textual modal local features as sentences. They adopted a weighted cross-modal attention mechanism similar to GLoRIA to achieve cross-modal alignment of local features. 

Similar to the general domain, Medical VLP faces significant interference from false-negative samples during training. KOBO \cite{chen2023knowledge} and SAT \cite{liu2023improving} from the perspectives of constructing medical knowledge graphs and leveraging existing medical LLM models, expand the similarity labels of image reports in the existing dataset by computing report similarities. This process mitigates the impact of false-negative samples on the model.
On the other hand, REFERS \cite{zhou2022generalized} extends the existing cross-modal alignment task by introducing an additional report generation task, enhancing the semantic co-occurrence learning between images and reports. CXR-CLIP \cite{you2023cxr} introduces an intra-modal alignment task, strengthening the intra-modal semantic consistency learning. 

While medical VLP methods have achieved success through cross-modal contrastive learning, they encounter limitations in capturing crucial local and fine-grained modal features necessary for medical cross-modal retrieval tasks. To tackle this, this paper introduces the mask reconstruction proxy task and proposes multi-task fusion frameworks. These enhancements maintain the advantages of cross-modal contrastive learning while enhancing the extraction of essential modal features, consequently improving medical cross-modal retrieval task performance. The primary focus of this paper is on effectively modifying VLP to further enhance its performance in medical cross-modal retrieval tasks.

\section{Masked Contrastive Reconstruction}
The existing VLP framework that combines cross-modal contrastive and masked reconstruction commonly inputs unmasked and masked data pair $\left(x, \hat{x}\right)\text{,}$ through the shared modality encoder $E$ to generate unmasked features $f$ and masked features $\hat{f}$. Subsequently, it uses $f$ for cross-modal contrastive learning and $\hat{f}$ for masked reconstruction learning. 
In this VLP framework, there are significant differences between the inputs of these proxy tasks. This leads to task competition and information interference during training, severely limiting the learning of intra-modal features and cross-modal common semantic features, as shown in \cite{shrestha2023medical}.
For this, we propose an efficient VLP framework called \textbf{M}asked \textbf{C}ontrastive and \textbf{R}econstruction (MCR), which utilizes $\hat{f}$ for both cross-modal contrastive learning and masked reconstruction. This contributes to enhancing cross-modal alignment and improving the performance of cross-modal retrieval tasks. Without differing data distributions between the inputs of both proxy tasks, the MCR framework integrates these tasks more effectively. Moreover, as it no requires unmasked data input, MCR offers advantages in reducing memory usage and enhancing training speed compared to existing frameworks. For example, our MCR framework conserves 75\% of GPU memory and reduces time consumption by 50\% during training. {\color{black}{Furthermore, to ensure practicality in large-scale cross-modal retrieval applications, we deliberately maintained the independence of different modal feature extraction processes and avoided the use of multi-modal feature fusion modules.}} 

\subsection{Masked Feature Representation}

In this paper, we use a dual-stream VLP architecture to extract modal features from both chest X-ray images and medical reports. For better feature alignment, we employ both VIT and Bert as feature encoders, both of which are based on the transformer framework. 

\textbf{Masked Visual Features:} Before inputting it into VIT, we divide a given chest X-ray image ${x_i^v \in \mathbb{R}^{H \times W \times C}}$ into $N$ non-overlapping patches $p_i = \left\{\ p_i^1,\ p_i^2,\ \ldots,\ p_i^N\ \right\}$, $p_i^n \in \mathbb{R}^{P \times P \times C}$. We employed the masking approach similar to MAE, randomly selecting patches in $p_i$ using a fixed masking rate. The remaining patch set $\hat{p}_i=\left\{p_i^k\ |\ k \notin M_i^v \right\}$ was used for extracting masked visual features. Here, $M_i^v$ represents the set of randomly masked indices. While MAE claims that a 75\% masking rate achieves optimal performance. In cross-modal tasks, an excessively large image masking proportion will exacerbate the difficulty of contrastive learning tasks, thereby impacting the alignment space learning. Hence, in this paper, we set the image masking rate to 50\%. Subsequently, we perform flattening and linear mapping operations on $\hat{p}_i$ to transform it into masked visual embeddings, denoted as $\hat{v}_i=\left\{\ v_i^0,\ v_i^1,\ \ldots,\ v_i^{\ \hat{N}}\ \right\} \in \mathbb{R}^{(\hat{N}+1)\times d}$. Here, $v_i^0$ represents the learnable [CLS] token, and $\hat{N}$ indicates the number of patches after masking. Adding this to the learnable visual positional embeddings $v_p \in \mathbb{R}^{(\hat{N}+1)\times d}$ results in the final visual modal input, which is then fed into the visual feature extractor $E_v$ to obtain the masked visual features $\hat{f}_i^{\ v}$.

\textbf{Masked Text Features:} For the radiology report $x_i^r$, we utilize the WordPiece algorithm olite{schuster2012japanese} to decompose it into a set of subword tokens denoted as $w_i = \left\{w_i^1,\ w_i^2,\ \ldots,\ w_i^M\right\} \in \mathbb{R}^{M \times L} $, where $M$ represents the length of the subword set, and $L$ represents the number of subwords in the vocabulary. Using a fixed masking rate, we replace the subwords in $w_i$ with [MASK] to obtain the masked set $\hat{w}_i = \{w_i^k\ |\ w_i^k = [MASK]\ if\ k \in M_i^r\}$. Here, $M_i^r$ represents the set of randomly masked indices. Subsequently, the text embeddings $r_i=\left\{r_i^0,\ r_i^1,\ \ldots,\ r_i^M, \ r_i^{M+1}\right\} \in \mathbb{R}^{(M+2) \times d}$ are retrieved from the vocabulary based on $\hat{w}_i$, with $r_i^0 \text{\ and\ } r_i^{M+1}$ respectively representing the learnable [CLS] and [SEP] tokens. Similarly, the learnable text positional embeddings $r_p \in \mathbb{R}^{(M+2)\times d}$ are added, obtaining the textual modality input, which is then fed into the text feature extractor $E_r$, ultimately producing the masked text features $\hat{f}_i^{\ r}$. 
{\color{black}{Unlike MaCo, which primarily focuses on masked contrastive reconstruction in the image modality and overlooks the text modality, our findings highlight the significance of the masked text modality. We advocate for the joint consideration of masked image and text modalities, as this approach yields the most effective results. Our subsequent ablation experiments highlight the necessity and superiority of conducting masked contrastive reconstruction simultaneously in both image and text modalities. }}

\subsection{Modality Alignment with MbA}
In general scenarios, cross-modal alignment commonly aggregates local features within different modal spaces to obtain global features. Then, these global features are mapped to a common space for alignment learning. We refer to this alignment strategy as Aggregate before Mapping (AbM). Unlike general image captions, radiology reports typically provide detailed descriptions of various regions within chest X-ray images, lacking common sequential or causal associations between these descriptions. Additionally, the regions describing lesions in images corresponding to the reports are often narrowly distributed and sparse. So, similar to general scenarios, aggregating local features within the original modality space often results in cross-modal consist information missing. 
Although this missing is inevitable during local feature aggregation, it is essential for medical cross-modal alignment.
As this missing occurs prior to mapping into a common alignment space, it leads to a suboptimal learned alignment space. Deep learning tend to focus on learning simpler content, which might exacerbate the learning of suboptimal alignment spaces, leading to a narrower and more one-sided focus on modal features. For this, we propose a method that initially maps all features from the original modality into a common space, aiming to minimize the loss of modal information. Subsequently, in this common space, these features are aggregated into global features for alignment learning. This approach enables a more refine alignment learning by preserving finer details. We refer to this alignment strategy as Mapping before Aggregate (MbA). We use $f_{v \to s}$ and $f_{r \to s}$ to denote the feature mapping functions from the visual and report spaces to the common space, where $s_v$ and $s_r$ represent the visual and textual features within the common space $s$, as shown below:
\begin{equation}
    \begin{aligned}
        & s_i^v = Agg(f_{v \to s}(\hat{f}_i^{\ v})\ ) ; \ \ s_i^r = Agg(f_{r \to s}(\hat{f}_i^{\ r})\ ).
    \end{aligned}
\end{equation}
The $Agg$ denotes an aggregation function, with MaxPool serving as its implementation in this paper. 
In the experimental section, this paper thoroughly evaluates the differences between AbM and MbA patterns both qualitatively and quantitatively, utilizing component ablation, embedding space analysis, group analysis, and retrieval results visualization. This evaluation confirms the effectiveness of the latter strategy.

\subsection{Masked Contrastive Reconstruction}
\textbf{Masked Contrastive Learning}: We employ the cross-modal alignment loss $L_{vrc}$ to achieve the goal of learning a common semantic space. It comprises two asymmetric sub-functions, denoted as $L^{\ s^v \to s^r}$ and $L^{\ s^r \to s^v}$:
\begin{equation}
    \small
    \begin{aligned}
        & \mathcal{L}_{vrc}(s^v, s^r) = \lambda_v L^{\ s^v\to s^r} + \lambda_r L^{\ s^r\to s^v}, \\
        & L^{\ s^v\to s^r} = -\sum_{k=1}^{B} log \left ( \frac{\exp( <\parallel{s_{k}^v}\parallel, \parallel s_{k}^r \parallel> / \tau)}{\sum_{j=1}^{B}\exp(<\parallel{s_{k}^v}\parallel, \parallel
        s_{j}^r \parallel> / \tau)}\right), \\
        & L^{\ s^r \to s^v} = -\sum_{k=1}^{B} log \left ( \frac{\exp( <\parallel{s_{k}^r}\parallel, \parallel s_{k}^v \parallel> / \tau)}{\sum_{j=1}^{B}\exp(<\parallel{s_{k}^r}\parallel, \parallel
        s_{j}^v \parallel> / \tau)}\right).
    \end{aligned}
\end{equation}
Here, $B$ represents the batch size, $\tau$ denotes the learnable temperature coefficient, $\parallel \parallel$ indicates feature normalization.  $\lambda_v$ and $\lambda_r$ stand for the weights of the asymmetric loss function, following the ConVIRT, they are set as $\lambda_v:\lambda_r=0.75:0.25$.

\begin{table*}[!t]
\normalsize
\caption{The performance comparison of different cross-modal retrieval methods on the MIMIC-CXR dataset. VLP, and MVLP represents Vision-Language Pre-training, and Medical Vision-Language Pre-training, respectively. MMU and CMA represent Multi-Modal Fusion and Cross-Modal Alignment, respectively.}
\centering
\begin{tabular}{cc|ccc|ccc}
\hline
\multicolumn{2}{c|}{}                                                    & \multicolumn{3}{c|}{I $\to$ R}                                          & \multicolumn{3}{c}{R $\to$ I}                                           \\ \hline
\multicolumn{1}{c|}{Method}                              & Category      & \multicolumn{1}{c}{Recall@1} & \multicolumn{1}{c}{Recall@5} & Recall@10 & \multicolumn{1}{c}{Recall@1} & \multicolumn{1}{c}{Recall@5} & Recall@10 \\ \hline
\multicolumn{1}{c|}{ConVIRT \cite{zhang2022contrastive}} & MVLP,\ CMA    & \multicolumn{1}{c}{6.765\%}  & \multicolumn{1}{c}{22.421\%} & 32.219\%  & \multicolumn{1}{c}{6.647\%}  & \multicolumn{1}{c}{16.151\%} & 23.832\%  \\
\multicolumn{1}{c|}{REFERS \cite{zhou2022generalized}}   & MVLP,\ CMA    & \multicolumn{1}{c}{8.294\%}  & \multicolumn{1}{c}{25.765\%} & 36.910\%  & \multicolumn{1}{c}{8.607\%}  & \multicolumn{1}{c}{18.929\%} & 28.755\%  \\
\multicolumn{1}{c|}{LoVT \cite{muller2022joint}}         & MVLP,\ CMA    & \multicolumn{1}{c}{10.601\%} & \multicolumn{1}{c}{30.715\%} & 42.017\%  & \multicolumn{1}{c}{11.504\%} & \multicolumn{1}{c}{24.855\%} & 36.067\%  \\
\multicolumn{1}{c|}{DiVE \cite{kim2023improving}}        & CMA      & \multicolumn{1}{c}{12.960\%} & \multicolumn{1}{c}{35.070\%} & 48.419\%  & \multicolumn{1}{c}{13.294\%} & \multicolumn{1}{c}{26.993\%} & 38.780\%  \\
\multicolumn{1}{c|}{CLIP \cite{radford2021learning}}     & VLP,\ CMA     & \multicolumn{1}{c}{14.256\%} & \multicolumn{1}{c}{36.392\%} & 48.263\%  & \multicolumn{1}{c}{14.700\%} & \multicolumn{1}{c}{29.342\%} & 40.160\%  \\
\multicolumn{1}{c|}{BLIP2 \cite{li2023blip}}             & VLP,\ CMA     & \multicolumn{1}{c}{17.159\%} & \multicolumn{1}{c}{41.498\%} & 53.447\%  & \multicolumn{1}{c}{18.875\%} & \multicolumn{1}{c}{34.974\%} & 47.657\%           \\
\multicolumn{1}{c|}{M3AE \cite{chen2022multi}}           & MVLP,\ MMU    & \multicolumn{1}{c}{15.267\%} & \multicolumn{1}{c}{42.898\%} & 56.739\%  & \multicolumn{1}{c}{17.171\%} & \multicolumn{1}{c}{36.153\%} & 50.135\%  \\ 
\multicolumn{1}{c|}{MaskCLIP \cite{dong2023maskclip}}    & MVLP,\ CMA   & \multicolumn{1}{c}{18.170\%} & \multicolumn{1}{c}{43.339\%} & 55.962\%  & \multicolumn{1}{c}{19.301\%} & \multicolumn{1}{c}{36.531\%} & {49.058\%} \\
\multicolumn{1}{c|}{CXR-CLIP \cite{you2023cxr}}          & MVLP,\ CMA   & \multicolumn{1}{c}{20.736\%} & \multicolumn{1}{c}{45.853\%} & 58.968\%  & \multicolumn{1}{c}{22.985\%} & \multicolumn{1}{c}{40.477\%} & 52.484\%  \\ \hline
\multicolumn{1}{c|}{MCR + MbA}                           & MVLP,\ CMA    & \multicolumn{1}{c}{24.598\%} & \multicolumn{1}{c}{52.281\%} & 65.241\%  & \multicolumn{1}{c}{27.354\%} & \multicolumn{1}{c}{45.709\%} & 58.758\%  \\ \hline
\end{tabular}
\label{mimic_result}
\end{table*}

\textbf{Masked Reconstruction}: We denote $D_v$ as the visual feature decoder. Following MAE, we input masked visual features $\hat{f}_i^{\ v}$ along with additional mask tokens. The mask token is a learnable vector used to identify the masked image blocks. The $D_v$ ultimately outputs a pixel value vector for the masked blocks, reconstructing the masked image blocks through transformation and recombination operations. We use the mean squared error function (MSE) to compute the normalized pixel-wise difference between the original image patches $p_i^k$ and reconstructed image patches $\tilde{p}_i^k$, serving as the specific implementation of the visual masked reconstruction loss $L_{mim}$:
\begin{equation}
    \mathcal{L}_{mim} = \frac{1}{|M_i^v|}\sum_{k \in M_i^v}(p_i^k - \tilde{p}_i^k)^2.
\end{equation}

Similarly, we denote $D_r$ as the text feature decoder, taking masked text features as input to predict the original text representation corresponding to the [MASK] token, $\tilde{w}_i^k$ represent the predict output. We employ multi-label cross-entropy as the specific implementation of the text masked reconstruction loss $L_{mrm}$:
\begin{equation}
    \mathcal{L}_{mlm} = \frac{1}{|M_i^r|}\sum_{k\in M_i^r}CE(w_i^k, \tilde{w}_i^k).
\end{equation}

In conclusion, the final train objective function in this paper can be represented as:
\begin{equation}
    \mathcal{L} = \lambda_{vrc} \mathcal{L}_{vrc} + \lambda_{mim} \mathcal{L}_{mim} + \lambda_{mrm} \mathcal{L}_{mrm},
    \label{total_loss}
\end{equation}
where $\lambda_{vrc}$, $\lambda_{mim}$, and $\lambda_{mrm}$ represent the weight coefficients of different loss functions. The pipeline and the composition of loss functions for our method are depicted in Figure \ref{homepage}.

\section{Experiment}
\label{sec:experiment}
In this section, we first introduce the datasets and the details of experimental implementation. Subsequently, we introduce the evaluation metrics used in the experiments and compare them with previous state-of-the-art studies. Finally, we conduct comprehensive ablation experiments to validate the effectiveness of our method.

\begin{figure*}[!t]

    \centering
    \subfigure{
        \includegraphics[width=4.2cm]{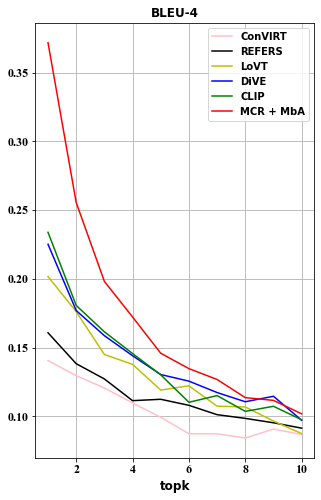}
    }
    \subfigure{
        \includegraphics[width=4.2cm]{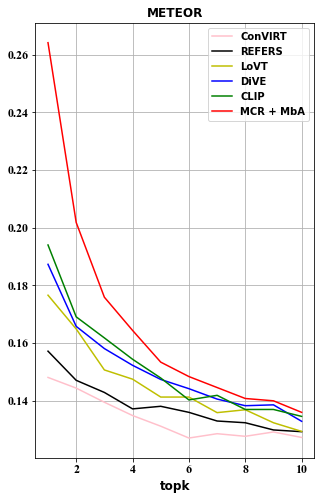}
    }
    \subfigure{
        \includegraphics[width=4.2cm]{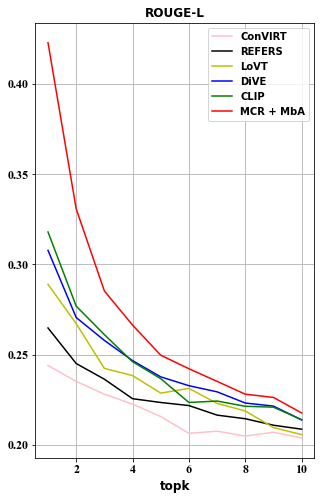}
    }
    \subfigure{
        \includegraphics[width=4.2cm]{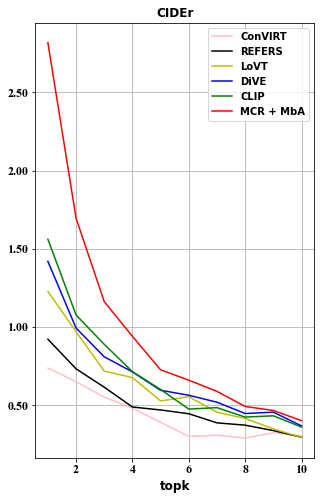}
    }
        \subfigure{
        \includegraphics[width=4.2cm]{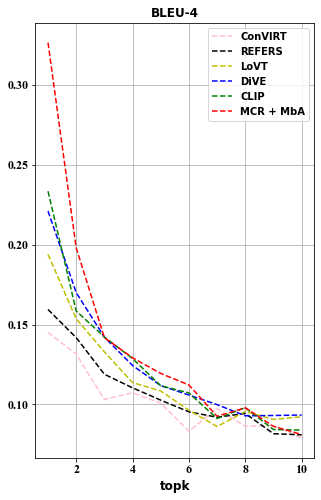}
    }
    \subfigure{
        \includegraphics[width=4.2cm]{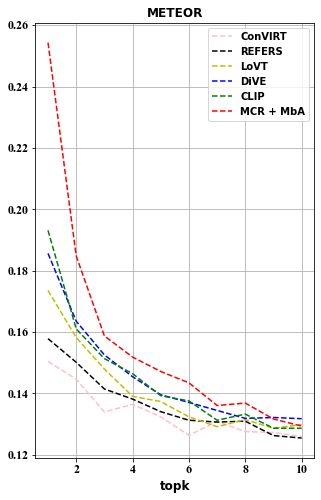}
    }
    \subfigure{
        \includegraphics[width=4.2cm]{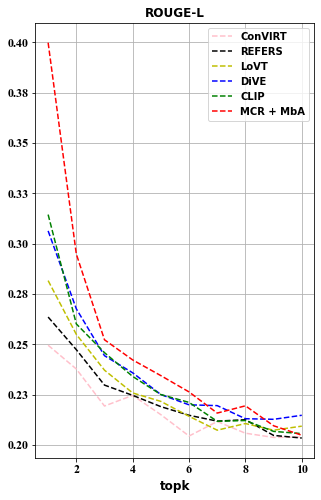}
    }
    \subfigure{
        \includegraphics[width=4.2cm]{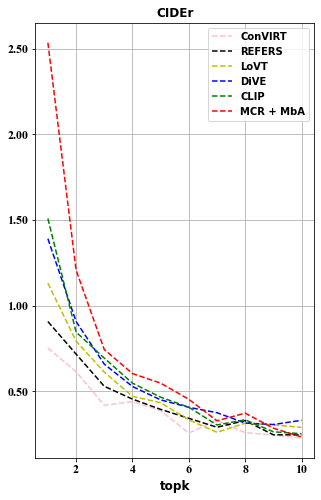}
    }
\caption{The TopK (K $\in$ [1, 10]) retrieval results in terms of NLG metrics (BLEU-4, METEOR, ROUGE-L, and CIDEr) for different cross-modal retrieval methods. Dashed and solid lines denote the subtask I $\to$ R and the subtask R $\to$ I, respectively.}
\label{nlg}
\end{figure*}

\subsection{Dataset}
MIMIC-CXR (v2.0.0) \cite{johnson2019mimic}  is the largest openly available chest X-ray image dataset to date, inclusive of data from 65,379 patients between 2011 and 2016. It comprises 377,100 chest X-ray images from different views and 227,835 semi-structured free-text radiology reports. Our research only focuses on the \textbf{findings} section of the reports, excluding samples without findings or with text lengths less than 3. Unlike previous studies, we maintained images from all views, including scenarios where multiple images from different views corresponded to a single report. Following the official data protocol, we partitioned the filtered dataset, resulting in the training set of 270,790 sample pairs, the validation set of 2,130 sample pairs, and the test set of 3,858 sample pairs.

\subsection{Implementation Details}
We employed pre-trained VIT-B16 \cite{dosovitskiy2020vit} as the visual encoder $E_v$ and pre-trained BioClinicalBERT \cite{alsentzer2019publicly} as the text encoder $E_r$ and tokenizer. In the dataset preprocessing stage, for every image, we resize the shorter side to 512 pixels while maintaining the unchanged aspect ratio. During training, we use RandomResizedCrop to resize images to 224 $\times$ 224, while during inference, we employ Resize to maintain the same dimensions. The learning rates for $E_v$ and $E_r$ were set to 1e-4, while the remaining network model parameters were set to 3e-4. 
AdamW was used as the training optimizer, with the batch size of 256, and the total training of 100 epochs. Moreover, we use mixed precision training technique to accelerate the training process. The learning rate started with a warm-up from zero for the first 10 epochs, followed by a cosine decay for the remaining epochs. Based on the experimental results, we set $\lambda_{vrc}$\ :\ $\lambda_{mim}$\ :\ $\lambda_{mrm}$ = 0.1\ :\ 1.0\ :\ 1.0.

\subsection{Comparisons with Previous Studies}
CMIIR typically involves two scenarios: images query reports (I$\to$R) and reports query images (R$\to$I). In our experimental setup, each image in I$\to$R corresponds uniquely to a single report, whereas in R$\to$I, each report corresponds to at least one view image. We employ Recall@K as the final evaluation metric, indicating the percentage of matches with the query object in the topK retrieval results. In this study, we set K $\in \{1, 5, 10\}$ to comprehensively assess the performance of various cross-modal retrieval algorithms. In our experimental setup, a single report might correspond to multiple images. Hence, when calculating the recall for subtask R$\to$I, it's crucial to note that the total recall count always equals the smaller value between the topK and the number of images corresponding to the report.

In the same experimental setup, we replicated several medical VLP models (ConVIRT, REFERS, LoVT, M3AE, and CXR-CLIP) using open-source code. Furthermore, we implemented two general domain VLP models (CLIP and BLIP2), along with the state-of-the-art non-VLP cross-modal retrieval algorithm, DiVE. Additionally, we've implemented MaskCLIP, respecting the framework by employing both masked and unmasked dual inputs. Our approach was compared with these models on the MIMIC-CXR across subtasks I$\to$R and R$\to$I, and the results are shown in Table \ref{mimic_result}. 

{\color{black}{Firstly, the method proposed in this paper significantly outperforms existing methods across all cross-modal retrieval}} {\color{black}{subtasks. Secondly, involving masked reconstruction tasks has demonstrated significant advantages in cross-modal retrieval compared to those without it. Notably, CXR-CLIP has added advanced image and text self-supervised contrastive proxy tasks to enhance intra-modal feature extraction, surpassing a series of methods involving mask reconstruction in cross-modal retrieval tasks. This success is attributed to the unified input among multiple proxy tasks within CXR-CLIP, eliminating multi-task information interference and competitive phenomena. This validates the importance of unifying inputs across different tasks for multi-task training, further confirming the rationale and necessity of our MCR framework.}}

\begin{table*}[!t]
\normalsize
\caption{The Ablation study for different components of our method on the MIMIC-CXR Dataset. Con, MC, MIM, MRM respectively represent standard feature contrastive, masked feature contrastive, masked image reconstruction, and masked report reconstruction components, respectively. $AbM$ serves as the default component if not explicitly specified.}
\centering
\begin{tabular}{cc|ccc|ccc} \hline                     
\multicolumn{2}{c|}{}         & \multicolumn{3}{c|}{I $\to$ R} & \multicolumn{3}{c}{R $\to$ I} \\ \hline
\multicolumn{1}{c|}{Index} &{Component} & \multicolumn{1}{c|}{Recall@1}  & \multicolumn{1}{c|}{Recall@5} & Recall@10 & \multicolumn{1}{c|}{Recall@1} & \multicolumn{1}{c|}{Recall@5} & Recall@10 \\ \hline
\multicolumn{1}{c|}{a)} & {MC + MIM}             & \multicolumn{1}{c|}{14.256\%} & \multicolumn{1}{c|}{37.506\%} & 48.808\%  & \multicolumn{1}{c|}{15.296\%} & \multicolumn{1}{c|}{30.238\%} & {40.641\%} \\ 
\multicolumn{1}{c|}{b)} & {MC + MRM}             & \multicolumn{1}{c|}{18.403\%} & \multicolumn{1}{c|}{43.546\%} & 55.780\%  & \multicolumn{1}{c|}{18.918\%} & \multicolumn{1}{c|}{34.459\%} & {46.935\%} \\  
\multicolumn{1}{c|}{c)} & {Con + MIM + MRM (MaskCLIP)} & \multicolumn{1}{c|}{18.170\%} & \multicolumn{1}{c|}{43.339\%} & 55.962\%  & \multicolumn{1}{c|}{19.301\%} & \multicolumn{1}{c|}{36.531\%} & {49.058\%} \\ 
\multicolumn{1}{c|}{d)} & {MC + MIM + MRM (MCR)}       & \multicolumn{1}{c|}{22.576\%} & \multicolumn{1}{c|}{49.352\%} & 61.457\%  & \multicolumn{1}{c|}{24.286\%} & \multicolumn{1}{c|}{42.414\%} & 55.392\%   \\
\multicolumn{1}{c|}{e)} & {Con + MIM + MRM + MbA} & \multicolumn{1}{c|}{22.317\%} & \multicolumn{1}{c|}{48.548\%} & 59.824\%  & \multicolumn{1}{c|}{24.585\%} & \multicolumn{1}{c|}{41.602\%} & 52.751\%   \\ 
\multicolumn{1}{c|}{f)} & {MC + MIM + MRM + MbA} & \multicolumn{1}{c|}{24.598\%} & \multicolumn{1}{c|}{52.281\%} & 65.241\%  & \multicolumn{1}{c|}{27.354\%} & \multicolumn{1}{c|}{45.709\%} & 58.758\%  \\ \hline
\end{tabular}
\label{ab_result}
\end{table*}
\begin{figure*}[!t]

    \centering
    \subfigure[AbM]{
        \label{abm}
        \includegraphics[width=8.6cm]{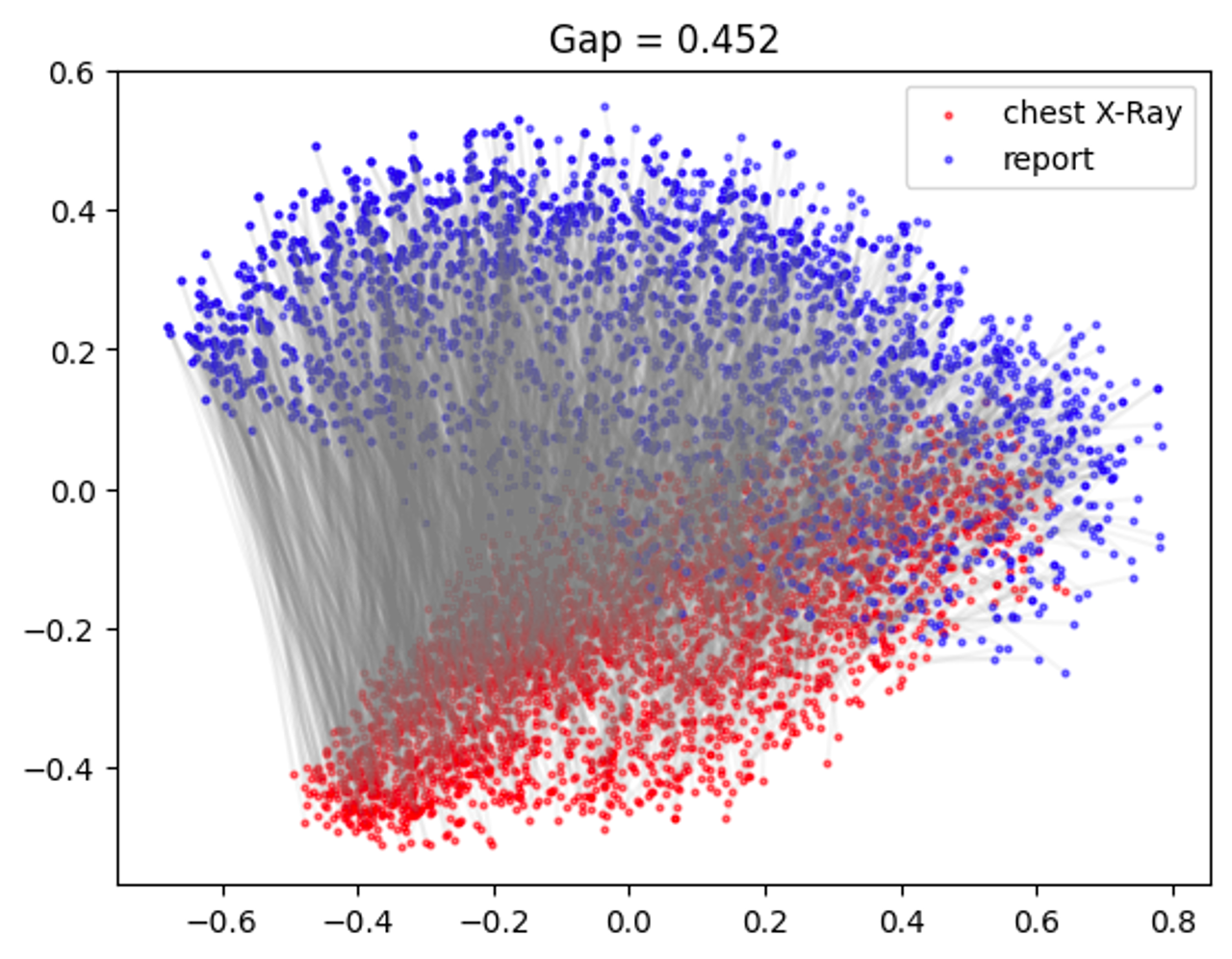}
    }
    \subfigure[MbA]{
        \label{mba}
        \includegraphics[width=8.6cm]{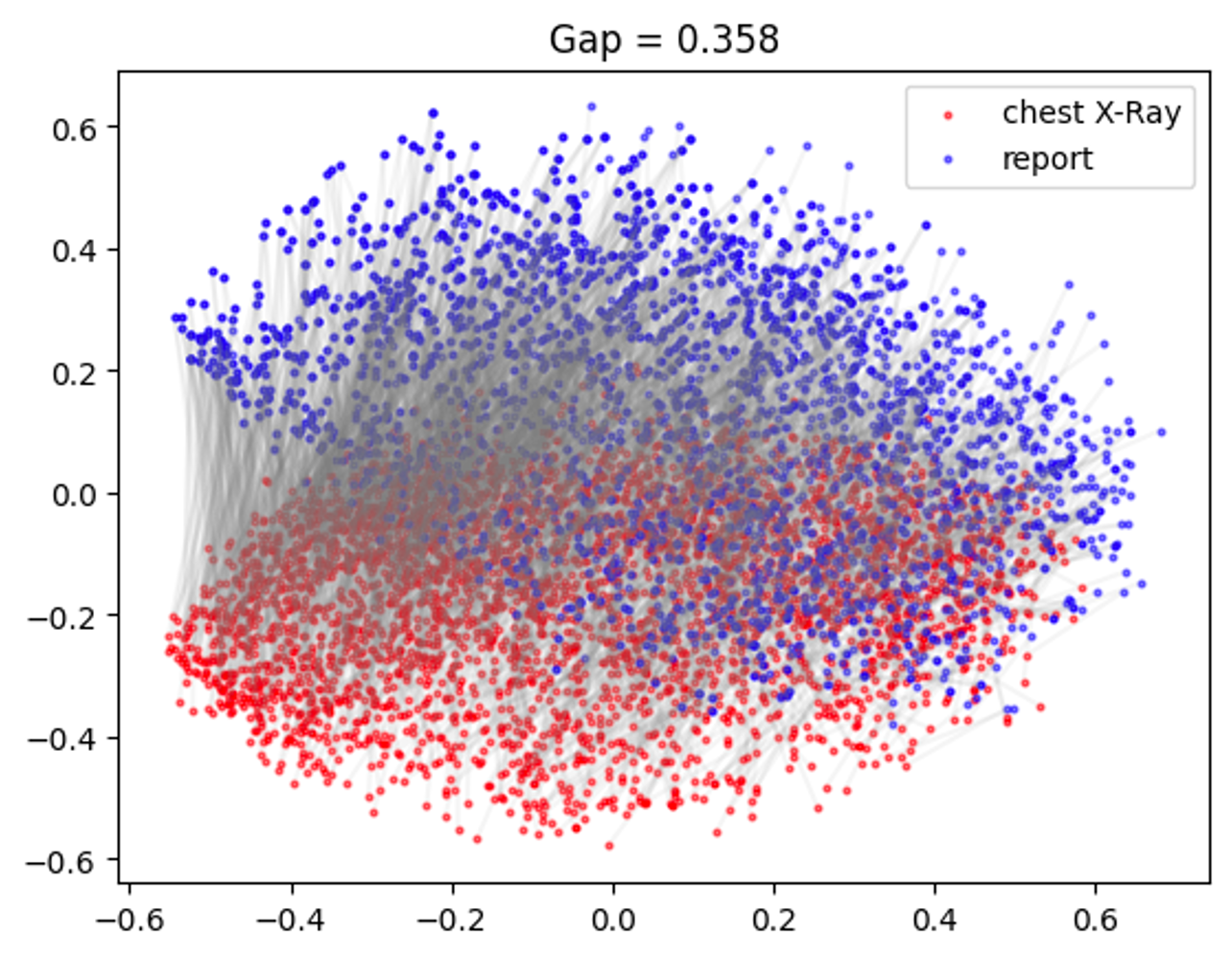}
    }
\caption{Visualization of the modality gap for our MCR with AbM (a) and MbA (b). The first two principal components of the validation set data are shown using t-SNE, where points in red and blue denote the embedded chest X-ray images and corresponding reports, respectively.}
\label{tsne}
\end{figure*}

In our comparative experiments, M3AE stands as the sole example of multi-modal fusion retrieval. Under the same inference environment and scale, M3AE takes approximately \textbf{28 hours} to complete all retrievals, whereas all cross-modal contrastive retrieval methods only require \textbf{35 to 60 seconds}. This clearly demonstrates the substantial computational cost incurred by multi-modal information fusion, a cost that fundamentally cannot meet the practical demands of large-scale cross-modal retrieval applications.

Medical cross-modal retrieval faces the challenge of false negatives, where image reports, although not in positive pairs, still express similarity information. For instance, in the MIMIC-CXR dataset, 'no abnormality' samples exhibit similarities between different pairs. We used evaluation metrics derived from medical report generation tasks to assess the accuracy of report descriptions for the topk 1-10 retrieval results. Evaluation metrics include BLEU-4 \cite{papineni2002bleu}, ROUGE-L \cite{lin2004rouge}, METEOR \cite{banerjee2005meteor}, and CIDEr \cite{Vedantam_2015_CVPR}, providing a comprehensive assessment of various cross-modal retrieval algorithms. From Figure \ref{nlg}, it is evident that our method significantly outperforms previous cross-modal retrieval methods across both sub-tasks and all report generation metrics. It consistently maintains a descending trend from top1 to top10, unlike other methods that display fluctuations. This trend persists due to the capability of our method to finely rank results based on semantic similarity, ensuring higher similarity instances are positioned at the forefront while lower similarity instances are placed towards the end.

\subsection{Ablation Experiments}

\begin{table}[t]
\caption{Ablation experiments on masking ratios (MR) of the report.}
\centering
\normalsize
\begin{tabular}{cc|ccc}
\hline
\multicolumn{1}{c|}{}                           & \multicolumn{1}{c|}{MR}   & Recall@1                     & Recall@5                     & Recall@10 \\ \hline
\multicolumn{1}{c|}{\multirow{3}{*}{I $\to$ R}} & \multicolumn{1}{c|}{\ 15\%\ } & \multicolumn{1}{c}{22.965\%} & \multicolumn{1}{c}{51.192\%} & 63.349\%  \\
\multicolumn{1}{c|}{}                           & \multicolumn{1}{c|}{\ 25\%\ } & \multicolumn{1}{c}{24.598\%} & \multicolumn{1}{c}{52.281\%} & 65.241\%  \\
\multicolumn{1}{c|}{}                           & \multicolumn{1}{c|}{\ 50\%\ } & \multicolumn{1}{c}{23.224\%} & \multicolumn{1}{c}{51.400\%} & 63.291\%  \\ \hline
\multicolumn{1}{c|}{\multirow{3}{*}{R $\to$ I}} & \multicolumn{1}{c|}{\ 15\%\ } & \multicolumn{1}{c}{25.394\%} & \multicolumn{1}{c}{44.126\%} & 55.951\%  \\
\multicolumn{1}{c|}{}                           & \multicolumn{1}{c|}{\ 25\%\ } & \multicolumn{1}{c}{27.354\%} & \multicolumn{1}{c}{45.709\%} & 58.758\%  \\
\multicolumn{1}{c|}{}                           & \multicolumn{1}{c|}{\ 50\%\ } & \multicolumn{1}{c}{25.479\%} & \multicolumn{1}{c}{43.600\%} & 56.951\%  \\ \hline
\end{tabular}
\label{masking}
\vspace{-10pt}
\end{table}

\begin{table}[t]
\caption{Comparison of GPU memory and training time required between MaskCLIP and MCR.}
\centering
\normalsize
\begin{tabular}{c|cc}
\hline
\multicolumn{1}{c|}{}          & \multicolumn{1}{c|}{Training Times}               & \multicolumn{1}{c}{GPU Memory}           \\ \hline
\multicolumn{1}{c|}{MaskCLIP}  & \multicolumn{1}{c|}{\ $\sim$ 55h 45min \ }     & \multicolumn{1}{c}{17,490MiB * 4GPUs}    \\ 
\multicolumn{1}{c|}{MCR}       & \multicolumn{1}{c|}{\ $\sim$ 24h 12min \ }     & \multicolumn{1}{c}{17,262MiB * 2GPUs}    \\ \hline
\end{tabular}
\label{masking}
\end{table}

\begin{table*}[]
\caption{The RECALL@K Results of Different Alignment strategies with Varied Report Sentence Lengths on the MIMIC-CXR Dataset. +$AbM$ and +$MbA$ represent ablation experiments d) and f), respectively. $\Delta$ represents the gain of +$AbM$ relative to +$MbA$. }
\centering
\normalsize
\begin{tabular}{cc|ccc|ccc}
\hline
                                           &           & \multicolumn{3}{c|}{I $\to$ R} & \multicolumn{3}{c}{R $\to$ I} \\ \hline
\multicolumn{1}{c|}{Sencetens Number}      & Recall@K    & + AbM        & + MbA     & $\Delta$      & + AbM         & + MbA         & $\Delta$           \\ \hline
\multicolumn{1}{c|}{\multirow{3}{*}{1 $\sim$ 5}}  & Recall@1  & 10.920\%   & 11.288\%  & + 0.368\%     & 14.058\%    & 13.462\%    & - 0.596\%  \\
\multicolumn{1}{c|}{}                      & Recall@5  & 28.658\%   & 30.016\%  & + 1.358\%     & 26.156\%    & 26.936\%    & + 0.780\%  \\
\multicolumn{1}{c|}{}                      & Recall@10 & 40.124\%   & 43.896\%  & + 3.772\%     & 36.748\%    & 41.766\%    & + 5.018\%  \\ \hline
\multicolumn{1}{c|}{\multirow{3}{*}{6 $\sim$ 10}} & Recall@1  & 31.732\%   & 34.354\%  & + 2.622\%     & 30.916\%    & 33.766\%    & + 2.850\%  \\
\multicolumn{1}{c|}{}                      & Recall@5  & 60.530\%   & 64.896\%  & + 4.366\%     & 51.730\%    & 54.370\%    & + 2.640\%  \\
\multicolumn{1}{c|}{}                      & Recall@10 & 71.612\%   & 76.808\%  & + 5.196\%     & 64.134\%    & 67.510\%    & + 3.376\%  \\ \hline
\multicolumn{1}{c|}{\multirow{3}{*}{11 or more}}  & Recall@1  & 41.048\%   & 66.166\%  & + 25.118\%    & 39.334\%    & 50.000\%    & + 10.666\% \\
\multicolumn{1}{c|}{}                      & Recall@5  & 67.692\%   & 85.192\%  & + 17.500\%    & 51.667\%    & 82.652\%    & + 30.985\% \\
\multicolumn{1}{c|}{}                      & Recall@10 & 91.310\%   & 89.380\%  & - 1.930\%     & 63.667\%    & 87.000\%    & + 23.333\% \\ \hline
\end{tabular}
\vspace{+5pt}
\label{performance}
\end{table*}
\begin{figure*}[!t]

    \centering
    \subfigure[Best matching case.]{
        \label{yes}
        \includegraphics[width=16cm]{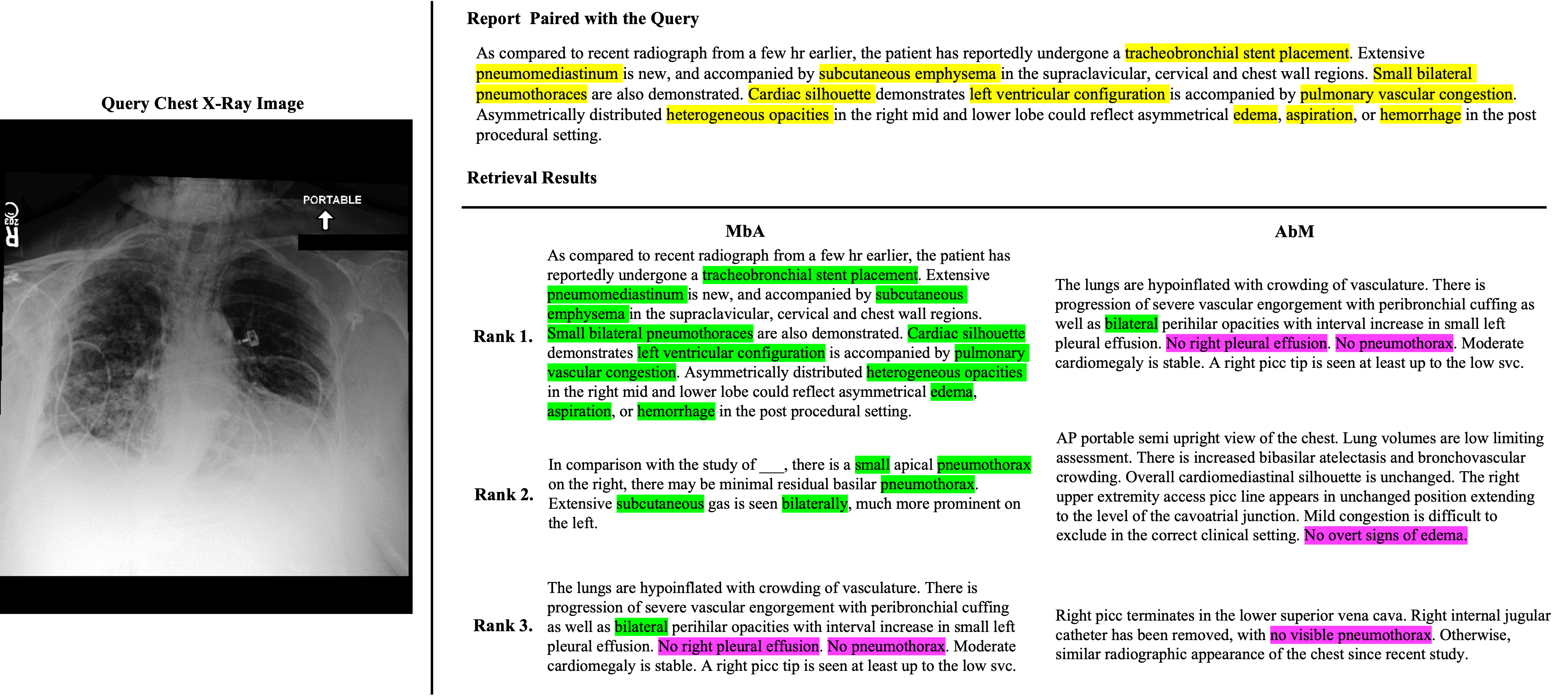}
    }
    \subfigure[Unsatisfactory matching case.]{
        \label{no}
        \includegraphics[width=16cm]{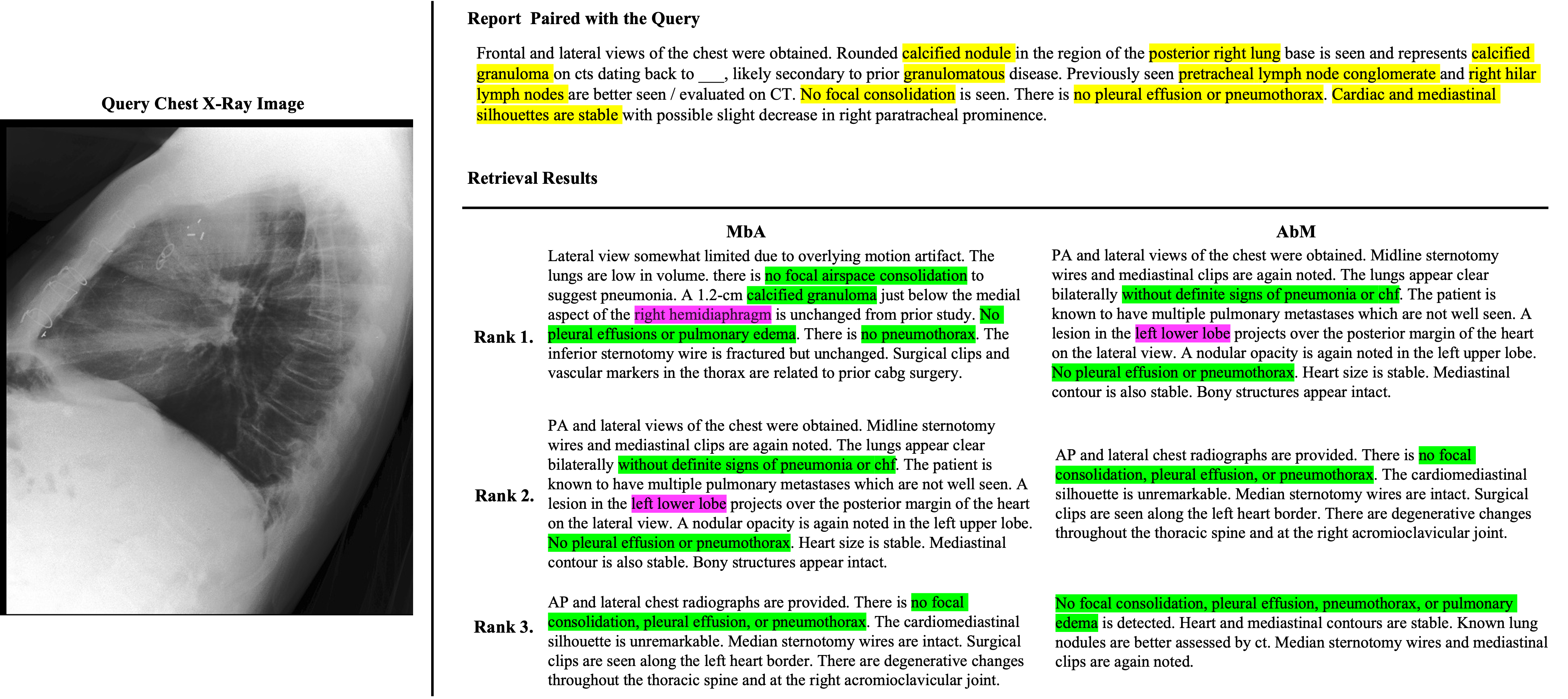}
    } \\
\caption{The Top-3 retrieval results using different alignment strategies. Text in yellow represents the medical semantics contained in the ground truth, text in green denotes the medical semantic content similar to the query sample, and text in pink represents the content irrelative to the query sample.}
\label{visual}
\end{figure*}

This section aims to validate the effectiveness of different component proposed in this paper. For clarity in description, C denotes standard contrastive, MC represents masked contrastive, MIM signifies visual masked reconstruction $L_{mim}$, and MRM indicates report masked reconstruction $L_{mrm}$.

\textbf{Objectives}: Experiments a), b), and d) in Table \ref{ab_result} illustrate the distinct contributions of difference objectives to the cross-modal retrieval task. The best performance is achieved by the combination of the three objective. Particularly, MRM significantly enhances retrieval performance compared to MIM. When only employing MIM, the results closely approximate those of pure CLIP, consistent with the findings in FLIP \cite{li2022scaling}. To further demonstrate the necessity of MRM, we conducted ablation experiments with different report masking rates, as shown in the Table \ref{masking}. It's evident that various masking rates are beneficial for the retrieval task, with the 25\% report masking rate used in this study yielding the best performance.

\textbf{Masked Contrastive}: Experiment pairs c), d) and e), f) in Table \ref{ab_result}, demonstrate that masked contrastive MC effectively enhances the performance of cross-modal retrieval tasks across alignment strategys. 
{\color{black}{It's worth emphasizing that experiment c) corresponds to MaskCLIP in Table \ref{mimic_result}. In both subtasks, the average increase in retrieval results is around 5\% after replacing with the Masked Contrastive. After unifying the inputs of contrastive learning and mask reconstruction using MCR, experiment d) demonstrates superior performance in cross-modal retrieval compared to CXR-CLIP. This validates the stronger feature extraction capability of the mask reconstruction proxy task over contrastive learning, aligning with the widely accepted consensus in the self-supervised learning domain.}}

We separately documented the training times and GPU memory usage for experiments c) and d), each run on the RTX 3090 devices, detailed in Table \ref{performance}. With only the single masked input needed, MCR demonstrates a 2 $\sim$ 3$\times$ improvement in both training times and GPU memory usage compared to MaskCLIP.

\textbf{Alignment strategy}: Experiment pairs c), e) and d), f) in Table \ref{ab_result}, demonstrate that MbA effectively enhances the performance of cross-modal retrieval tasks. In both subtasks, the average increase in retrieval results is around 4\% after replacing with the MbA alignment pattern.

Additionally, this paper uses t-SNE \cite{van2008visualizing} visualizations of the global feature embedding spaces of images and texts in the MIMIC-CXR test set under the alignment strategy of AbM and MbA, as illustrated in Figure \ref{tsne}. And calculated the Euclidean distance between different modal pairs defined as Gap in the t-SNE reduced space, which were 0.452 and 0.358, respectively. Under MbA pattern, there's a tighter embedding of image and text features in the common space, reducing heterogeneity between modalities. In contrast, while AbM pattern alignment shows overlapping feature distributions between modalities, noticeable disparities remain. These observations supports that MbA alignment achieves superior cross-modal alignment, consistent with the findings from quantitative retrieval experiments.

In radiology reports, a greater number of sentences indicates more complex modal information, making the aggregation of these local features into a complete global representation more challenging. To validate the effectiveness of the MbA alignment pattern in handling complex information within reports, we conducted the following validation experiment: Firstly, we segmented all reports in the test set using the NLTK tool \cite{bird2006nltk} and grouped them based on the number of sentences into 3 categories: 1$\sim$5, 6$\sim$10, and 11 or more. Subsequently, we analyzed the retrieval results from ablation experiments d) and f) within these groups, as detailed in Table \ref{ab_result}. The results revealed that, after applying MbA, the group with 11 or more exhibited the most substantial average gains (+17.612\%), followed by the 6-10 group (+3.508\%), whereas the 1-5 group exhibited nearly no gains (+1.783\%). This further indicates the ability of MbA to effectively handle complex information within reports by aggregating complete local features, reducing the loss of modal information, and achieving finer cross-modal alignment.

Moreover, we selected an best matching retrieval case (paired ranking 1st in the retrieval candidate list) and an unsatisfactory matching case (paired ranking not in topk10 of the retrieval candidate list), as shown in Figure \ref{visual} for visual qualitative analysis. 
In the visualization, yellow represents the medical semantics contained in the ground truth (annotated by professional physicians), green denotes medical semantic content similar to the query sample, and red represents content dissimilar to the query sample. 
We display the top three returned results under different alignment modes. 
It is evident that the MbA retrieval consistently encompasses more refine semantic content, covering multiple distinct medical pieces of information. 
For instance, in the failed case, all returned results from MbA contain information regarding \textbf{granulomatous} descriptions, while in AbM, all returned results focus solely on \textbf{no focal consolidation, pleural effusion, or pneumothorax}. 
Refer to the supplementary materials for further qualitative visualization experiments regarding retrieval results.

\section{Conclusion}
In this paper, we propose an efficient VLP framework named MCR, which solely utilizes masked data and unifies it as the uniform input across different tasks. This strengthens the connections between various tasks, reducing interference and competition among them, thereby enhancing the extraction of intra-modal features and cross-modal common semantic features.
Moreover, as it only requires masked data input, MCR also has advantages in reducing memory usage and enhancing training speed. 
For example, our MCR framework conserves 75\% of GPU memory and reduces time consumption by 50\% during training.
Additionally, we introduce a learning alignment paradigm termed MbA, tailored to the characteristics of chest X-ray images and radiology reports. This approach aims to minimize the loss of consistency information caused by challenging aggregation and achieve a more refined cross-modal alignment. Through qualitative and quantitative experiments on the MIMIC-CXR dataset, we validate the effectiveness of MCR and MbA, demonstrating their achievement of the SOTA performance in medical cross-modal retrieval tasks.

While our work unifies the data input for contrastive and reconstruction tasks, there remains an issue with the integration approach, lacking a sufficient level of coherence. We believe that a tighter integration approach will yield superior cross-modal alignment and consistency performance, and this will be our focus for the next phase of research.


\section*{References}
\begin{thebibliography}{00}\leftskip1pc
\setlength{\itemsep}{0bp}\setlength{\parskip}{0pt}\small
\bibitem{schuster2012japanese} Schuster M, Nakajima K. Japanese and korean voice search[C]. 2012 IEEE international conference on acoustics, speech and signal processing (ICASSP), 2012, 5149-5152.
\bibitem{zhang2023multi} Zhang K, Yang Y, Yu~J, et~al. Multi-task paired masking with alignment modeling for medical vision-language pre-training[J]. IEEE Transactions on Multimedia, 2023.
\bibitem{Ji_2023_CVPR} Ji~Y, Tu~R, Jiang J, et~al. Seeing what you miss: Vision-language pre-training with semantic completion learning[C]. Proceedings of the IEEE/CVF Conference on Computer Vision and Pattern Recognition (CVPR), 2023, 6789-6798.
\bibitem{dong2023maskclip} Dong X, Bao J, Zheng Y, et~al. Maskclip: Masked self-distillation advances contrastive language-image pretraining[C]. Proceedings of the IEEE/CVF Conference on Computer Vision and Pattern Recognition, 2023, 10995-11005.
\bibitem{hardoon2004canonical} Hardoon D.~R, Szedmak S, Shawe-Taylor J. Canonical correlation analysis: An overview with application to learning methods[J]. Neural computation, 2004, 16(12): 2639-2664.
\bibitem{rasiwasia2010new} Rasiwasia N, Pereira J.~C, Coviello E, et~al. A new approach to cross-modal multimedia retrieval[C]. Proceedings of the 18th ACM international conference on Multimedia, 2010, 251-260.
\bibitem{chen2012continuum} Chen Y, Wang L, Wang W, et~al. Continuum regression for cross-modal multimedia retrieval[C]. 2012 19th IEEE International Conference on Image Processing, 2012, 1949-1952.
\bibitem{andrew2013deep} Andrew G, Arora R, Bilmes J, et~al. Deep canonical correlation analysis[C]. International conference on machine learning, 2013, 1247-1255.
\bibitem{gong2014improving} Gong Y, Wang L, Hodosh M, et~al. Improving image-sentence embeddings using large weakly annotated photo collections[C]. Computer Vision--ECCV 2014: 13th European Conference, Zurich, Switzerland, September 6-12, 2014, Proceedings, Part IV 13, 2014, 529-545.
\bibitem{klein2014fisher} Klein B, Lev G, Sadeh G, et~al. Fisher vectors derived from hybrid gaussian-laplacian mixture models for image annotation[J]. arXiv preprint arXiv:1411.7399, 2014.
\bibitem{liu2023efficient} Liu C, Zhang Y, Wang H, et~al. Efficient token-guided image-text retrieval with consistent multimodal contrastive training[J]. IEEE Transactions on Image Processing, 2023.
\bibitem{frome2013devise} Frome A, Corrado G.~S, Shlens J, et~al. Devise: A deep visual-semantic embedding model[J]. Advances in neural information processing systems, 2013, 26.
\bibitem{zhang2016pl} Zhang L, Ma~B, Li~G, et~al. Pl-ranking: A novel ranking method for cross-modal retrieval[C]. Proceedings of the 24th ACM international conference on Multimedia, 2016, 1355-1364.
\bibitem{faghri2018vse++} Faghri F, Fleet D.~J, Kiros J.~R, et~al. Vse++: Improving visual-semantic embeddings with hard negatives[J]. , 2018.
\bibitem{chun2021pcme} Chun S, Oh~S.~J, Rezende R.~S.~D, et~al. Probabilistic embeddings for cross-modal retrieval[C]. Conference on Computer Vision and Pattern Recognition (CVPR), 2021, .
\bibitem{kim2023improving} Kim D, Kim N, Kwak S. Improving cross-modal retrieval with set of diverse embeddings[C]. Proceedings of the IEEE/CVF Conference on Computer Vision and Pattern Recognition, 2023, 23422-23431.
\bibitem{lee2018stacked} Lee K.-H, Chen X, Hua G, et~al. Stacked cross attention for image-text matching[C]. Proceedings of the European conference on computer vision (ECCV), 2018, 201-216.
\bibitem{zhang2020context} Zhang Q, Lei Z, Zhang Z, et~al. Context-aware attention network for image-text retrieval[C]. Proceedings of the IEEE/CVF conference on computer vision and pattern recognition, 2020, 3536-3545.
\bibitem{wang2019camp} Wang Z, Liu X, Li~H, et~al. Camp: Cross-modal adaptive message passing for text-image retrieval[C]. Proceedings of the IEEE/CVF international conference on computer vision, 2019, 5764-5773.
\bibitem{radford2021learning} Radford A, Kim J.~W, Hallacy C, et~al. Learning transferable visual models from natural language supervision[C]. International conference on machine learning, 2021, 8748-8763.
\bibitem{li2023blip} Li~J, Li~D, Savarese S, et~al. Blip-2: Bootstrapping language-image pre-training with frozen image encoders and large language models[J]. arXiv preprint arXiv:2301.12597, 2023.
\bibitem{liu2022image} Liu Z, Chen F, Xu~J, et~al. Image-text retrieval with cross-modal semantic importance consistency[J]. IEEE Transactions on Circuits and Systems for Video Technology, 2022.
\bibitem{lu2022cots} Lu~H, Fei N, Huo Y, et~al. Cots: Collaborative two-stream vision-language pre-training model for cross-modal retrieval[C]. Proceedings of the IEEE/CVF Conference on Computer Vision and Pattern Recognition, 2022, 15692-15701.
\bibitem{wang2023agree} Wang X, Li~L, Li~Z, et~al. Agree: Aligning cross-modal entities for image-text retrieval upon vision-language pre-trained models[C]. Proceedings of the Sixteenth ACM International Conference on Web Search and Data Mining, 2023, 456-464.
\bibitem{moon2022multi} Moon J.~H, Lee H, Shin W, et~al. Multi-modal understanding and generation for medical images and text via vision-language pre-training[J]. IEEE Journal of Biomedical and Health Informatics, 2022, 26(12): 6070-6080.
\bibitem{chen2022multi} Chen Z, Du~Y, Hu~J, et~al. Multi-modal masked autoencoders for medical vision-and-language pre-training[C]. International Conference on Medical Image Computing and Computer-Assisted Intervention, 2022, 679-689.
\bibitem{zhang2022contrastive} Zhang Y, Jiang H, Miura Y, et~al. Contrastive learning of medical visual representations from paired images and text[C]. Machine Learning for Healthcare Conference, 2022, 2-25.
\bibitem{huang2021gloria} Huang S.-C, Shen L, Lungren M.~P, et~al. Gloria: A multimodal global-local representation learning framework for label-efficient medical image recognition[C]. Proceedings of the IEEE/CVF International Conference on Computer Vision, 2021, 3942-3951.
\bibitem{muller2022joint} M{\"u}ller P, Kaissis G, Zou C, et~al. Joint learning of localized representations from medical images and reports[C]. European Conference on Computer Vision, 2022, 685-701.
\bibitem{Dawidowicz_2023_ICCV} Dawidowicz G, Hirsch E, Tal A. Limitr: Leveraging local information for medical image-text representation[C]. Proceedings of the IEEE/CVF International Conference on Computer Vision (ICCV), 2023, 21165-21173.
\bibitem{zhou2022generalized} Zhou H.-Y, Chen X, Zhang Y, et~al. Generalized radiograph representation learning via cross-supervision between images and free-text radiology reports[J]. Nature Machine Intelligence, 2022, 4(1): 32-40.
\bibitem{you2023cxr} You K, Gu~J, Ham J, et~al. Cxr-clip: Toward large scale chest x-ray language-image pre-training[C]. International Conference on Medical Image Computing and Computer-Assisted Intervention, 2023, 101-111.
\bibitem{park2023self} Park N, Kim W, Heo B, et~al. What do self-supervised vision transformers learn?[J]. arXiv preprint arXiv:2305.00729, 2023.
\bibitem{chen2023contrastive} Chen C, Zhong A, Wu~D, et~al. Contrastive masked image-text modeling for medical visual representation learning[C]. International Conference on Medical Image Computing and Computer-Assisted Intervention, 2023, 493-503.
\bibitem{huang2023enhancing} Huang W, Zhou H, Li~C, et~al. Enhancing representation in radiography-reports foundation model: A granular alignment algorithm using masked contrastive learning[J]. arXiv preprint arXiv:2309.05904, 2023.
\bibitem{johnson2019mimic} Johnson A.~E, Pollard T.~J, Berkowitz S.~J, et~al. Mimic-cxr, a de-identified publicly available database of chest radiographs with free-text reports[J]. Scientific data, 2019, 6(1): 317.
\bibitem{dosovitskiy2020vit} Dosovitskiy A, Beyer L, Kolesnikov A, et~al. An image is worth 16x16 words: Transformers for image recognition at scale[J]. ICLR, 2021.
\bibitem{alsentzer2019publicly} Alsentzer E, Murphy J.~R, Boag W, et~al. Publicly available clinical bert embeddings[J]. NAACL HLT 2019, 2019, 72.
\bibitem{zhu2023cross} Zhu L, Wang T, Li~F, et~al. Cross-modal retrieval: A systematic review of methods and future directions[J]. arXiv preprint arXiv:2308.14263, 2023.
\bibitem{Vedantam_2015_CVPR} Vedantam R, Zitnick C.~L, Parikh D. Cider: Consensus-based image description evaluation[C]. Proceedings of the IEEE Conference on Computer Vision and Pattern Recognition (CVPR), 2015, .
\bibitem{papineni2002bleu} Papineni K, Roukos S, Ward T, et~al. Bleu: a method for automatic evaluation of machine translation[C]. Proceedings of the 40th annual meeting of the Association for Computational Linguistics, 2002, 311-318.
\bibitem{lin2004rouge} Lin C.-Y. Rouge: A package for automatic evaluation of summaries[C]. Text summarization branches out, 2004, 74-81.
\bibitem{banerjee2005meteor} Banerjee S, Lavie A. Meteor: An automatic metric for mt evaluation with improved correlation with human judgments[C]. Proceedings of the acl workshop on intrinsic and extrinsic evaluation measures for machine translation and/or summarization, 2005, 65-72.
\bibitem{li2022scaling} Li~Y, Fan H, Hu~R, et~al. Scaling language-image pre-training via masking[C]. CVPR, 2023, .
\bibitem{bird2006nltk} Bird S. Nltk: the natural language toolkit[C]. Proceedings of the COLING/ACL 2006 Interactive Presentation Sessions, 2006, 69-72.
\bibitem{van2008visualizing} Maaten L.~Vder , Hinton G. Visualizing data using t-sne.[J]. Journal of machine learning research, 2008, 9(11).
\bibitem{chen2023knowledge} Chen, X., He, Y., Xue, C., Ge, R., Li, S. \& Yang, G. Knowledge Boosting: Rethinking Medical Contrastive Vision-Language Pre-training. {\em International Conference On Medical Image Computing And Computer-Assisted Intervention}. pp. 405-415 (2023)
\bibitem{liu2023improving} Liu, B., Lu, D., Wei, D., Wu, X., Wang, Y., Zhang, Y. \& Zheng, Y. Improving Medical Vision-Language Contrastive Pretraining with Semantics-aware Triage. {\em IEEE Transactions On Medical Imaging}. (2023)
\bibitem{oord2018representation} Oord, A., Li, Y. \& Vinyals, O. Representation learning with contrastive predictive coding. {\em ArXiv Preprint ArXiv:1807.03748}. (2018)

\bibitem{shrestha2023medical}Shrestha, P., Amgain, S., Khanal, B., Linte, C. \& Bhattarai, B. Medical Vision Language Pretraining: A survey. {\em ArXiv Preprint ArXiv:2312.06224}. (2023)

\end{thebibliography}
\end{document}